\pdfoutput=1
\documentclass[11pt]{article}
\usepackage[]{naacl2021}
\usepackage{times}
\usepackage{latexsym}
\usepackage[T1]{fontenc}

\usepackage[utf8]{inputenc}
\usepackage{microtype}
\usepackage{amsmath}
\usepackage{graphicx,accents}
\usepackage{esvect}
\usepackage{comment}
\usepackage{algorithm}
\usepackage{algorithmic}
\usepackage{array}
\newcolumntype{P}[1]{>{\centering\arraybackslash}p{#1}}

\usepackage{graphicx}
\usepackage{footmisc}
\usepackage[T1]{fontenc}
\usepackage[utf8]{inputenc}
\usepackage{microtype}

\makeatletter
\DeclareRobustCommand{\cev}[1]{%
  \mathpalette\do@cev{#1}%
}
\newcommand{\do@cev}[2]{%
  \fix@cev{#1}{+}%
  \reflectbox{$\m@th#1\vec{\reflectbox{$\fix@cev{#1}{-}\m@th#1#2\fix@cev{#1}{+}$}}$}%
  \fix@cev{#1}{-}%
}
\newcommand{\fix@cev}[2]{%
  \ifx#1\displaystyle
    \mkern#23mu
  \else
    \ifx#1\textstyle
      \mkern#23mu
    \else
      \ifx#1\scriptstyle
        \mkern#22mu
      \else
        \mkern#22mu
      \fi
    \fi
  \fi
}

\makeatother

\title{STRATA : Word Boundaries \& Phoneme Recognition From Continuous Urdu Speech  using Transfer Learning, Attention, \& Data Augmentation}
\title{STRATA: Word Boundaries \& Phoneme Recognition From Continuous Urdu Speech  using Transfer Learning, Attention, \& Data Augmentation}

\author{Saad Naeem \\
Department of Computer Science\\ (FAST-NUCES) \\
    Islamabad, Pakistan\\
  \texttt{saadnaeem.dev@gmail.com} \\\And
   Dr. Mirza Omer Beg \\ Department of Computer Science\\
 (FAST-NUCES) \\
    Islamabad, Pakistan\\
  \texttt{omer.beg@nu.edu.pk} \\}

\begin{document}
\maketitle
\begin{abstract}
Phoneme recognition is a largely unsolved problem in NLP, especially for low-resource languages like Urdu. The systems that try to extract the phonemes from audio speech require hand-labeled phonetic transcriptions. This requires expert linguists to annotate speech data with its relevant phonetic representation which is both an expensive and a tedious task. In this paper, we propose STRATA, a framework for supervised phoneme recognition that overcomes the data scarcity issue for low resource languages using a seq2seq neural architecture integrated with \textit{transfer learning}, \textit{attention mechanism}, and \textit{data augmentation}. STRATA employs transfer learning to reduce the network loss in half. It uses attention mechanism for word boundaries and frame alignment detection which further reduces the network loss by 4\% and is able to identify the word boundaries with 92.2\% accuracy. STRATA uses various data augmentation techniques to further reduce the loss by 1.5\% and is more robust towards new signals both in terms of generalization and accuracy. STRATA is able to achieve a Phoneme Error Rate of 16.5\% and improves upon the state of the art by 1.1\% for TIMIT dataset (English) and 11.5\% for CSaLT dataset (Urdu).
\end{abstract}

\section{Introduction}

\noindent Automated speech recognition is a complex problem to model and require large amounts of transcribed data. End-to-End deep neural network based approaches requires 6-7 thousand hours of transcribed data for reasonable performance. Gathering this much data is not possible for Under-Resourced languages like Urdu.

The authors of deepspeech \cite{hannun2014deep} trained their systems on 7380 hrs. of data with 36 thousand different speakers to achieve 16\% word error rate. Since Urdu is an under-resourced language, the amount of data that is publicly available to train an Urdu ASR is only 126 hours at max which is not enough to train an  End-to-End speech recognition system that can compete with the state of the art \citep{hannun2014deep} in high resource languages. As an alternate we look towards GMM HMM based models that were able to give  32.74\% word error rate \cite{ali2014complete} for Arabic  that is also an low resource language. The gaussian mixture models combined with hidden Markov models can be implemented using Kaldi toolkit \cite{povey2011kaldi} which has its own challenges among which, producing the phonetic transcriptions is the most tedious task that also requires domain experts to understand and annotate the speech with its corresponding phonetic representation. The amount of phonetically transcribed data that is publicly available for Urdu is only 1.15 hrs. and in order to train a  GMM-HMM model we require around one thousand hours of data \cite{kamath2019deep} that is phonetically transcribed to get 23\% word error rate which is just not possible to do manually. 

The major contributions of this paper involves training a Seq2Seq Model for identifying word boundaries as well as extracting the phonetic representations of those words from continuous audio speech by (1) Applying transfer learning from pre-trained models in English. (2) Using a Bidirectional-attention mechanism to identify the word boundaries and frame alignment of phonemes in those words. (3) Using data augmentation for audio speech to counter issues related to data scarcity and generalization to unseen signals during test time. (4) Improving on state of the art by 1.1\% for English TIMIT dataset and 11.5\% in accuracy improvement for CSaLT (Urdu) dataset. (5) STRATA framework is generalizable to a broad range of problems that deals with audio sequential data and can be incorporated without making too much modifications in the existing solutions. (6) Our approach can be used for isolating word in continuous speech using the word boundary objective which we will explore in our future work. To the best of our knowledge this approach is not pursued by any of the earlier works published in Under-Resourced category.

\section{STRATA: Seq2seq  TransfeR  AugmenT Attention } 
Word boundaries and Phoneme recognition task for Urdu was accomplished using a Seq2Seq model that was initialized with the pre-trained model weights trained on common speech voice data (Source Model) next, the Destination Model's Architecture was modified for the target domain (Phoneme recognition) by changing the number of classes and adding an attention layer \cite{moore2014joint}. Finally, the data was augmented to overcome data scarcity using various techniques and the final model was trained. All of the steps are discussed in detail in the subsequent sections. 


The high level overview of the approach can be summarized as taking a pre-trained seq2seq network that converts speech to text in English language and apply transfer learning. We then modify the network into outputting the word boundaries and classes that represent the phonemes, add an attention layer \cite{wang2018self} and train this network (after transferring the weights) on augmented data + 1.15 hours of readily available data. This network is then used to output the word boundaries and their phonetic representations from continuous audio speech. Figure 1 shows the starting point of our approach that are the MFCC features.
\subsubsection{MFCC Features}
\begin{figure}[htb]
\centerline{\includegraphics[width=0.4\textwidth]{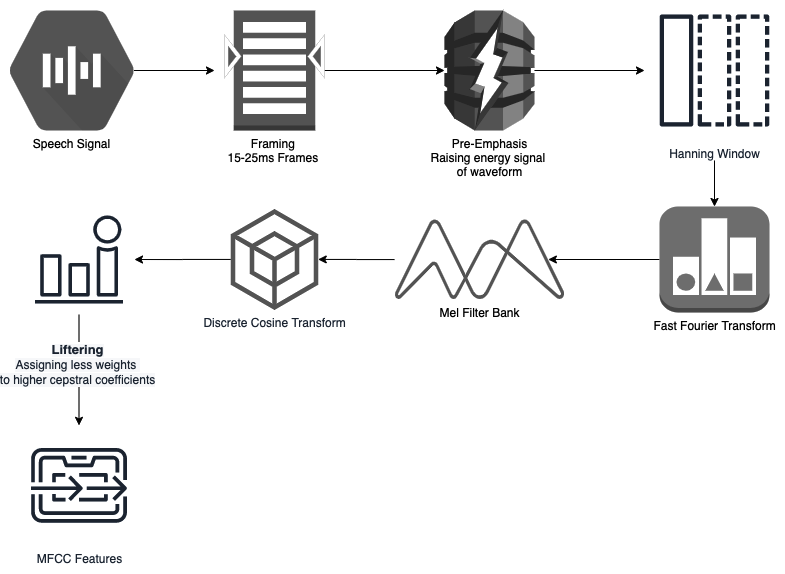}}
\caption{The final features that are used to train the network are MFCC features extracted from the original and augmented speech.}
\label{kmeans}
\end{figure}

\noindent The speech signal is first split into 10-30ms frames since this is an agreed upon time duration that can represent an individual sound unit \textbf{(Steve Young 1996)}. 
\noindent In pre-emphasis step first order differential is applied to the voiced sample to raise the energy level in speech waveform. Since Segmenting results in discontinuities among the frames Hanning window is used for smoothing the abrupt frame edges. Then Fast Fourier Transform is used to bring the signal from time domain to frequency domain \citep{kamath2019deep} and finally Audio pitch is calculated based on Mel Scale to bring the pitch to logarithmic scale rather than the linear scale \cite{paliwal1982performance}. The final MFCCs were calculated on 25 millisecond frames with steps size also equal to 25 millisecond i.e. no overlapping between the frames.

\subsubsection{Seq2Seq Architecture}
Applying Deep Neural Network for extracting the word boundaries and phonemes alleviates the need for explicit frame to phoneme alignment especially when the input are audio frames with much larger varying distribution.  The Neural Network Architecture used for extracting the phoneme sequences closely matches the Deep Speech  architecture shown in Figure 2. The inputs are the mel spectrograms of the audio with a window size of 25ms that flows through three fully connected layers followed by bi-directional Long Short Term Memory (LSTM) layer and again a single FC layer that feeds directly into the CTC Loss function.
\begin{figure}[]
\centerline{\includegraphics[width=0.4\textwidth]{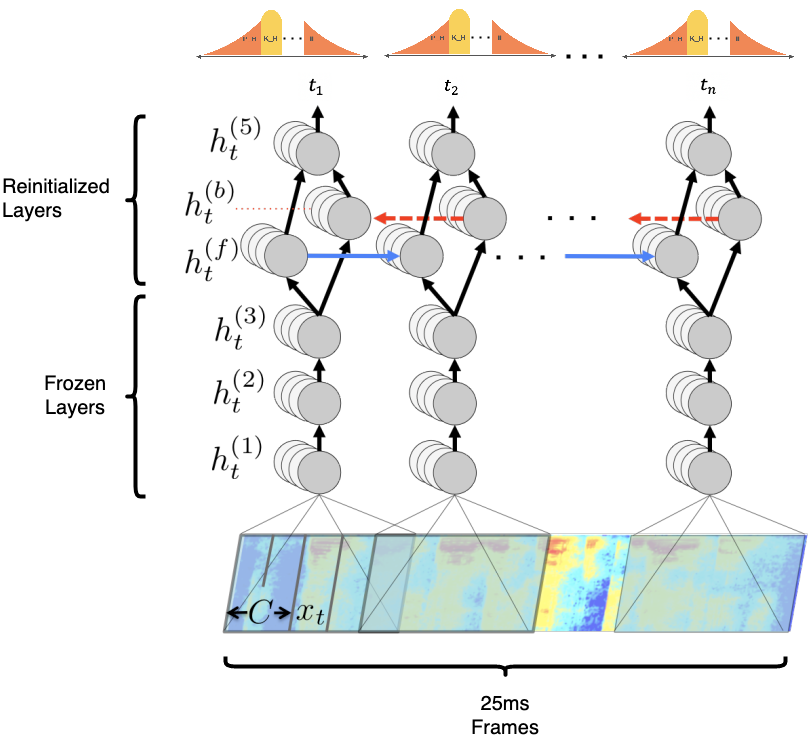}}

\caption{Deep Speech uses fixed length audio vectors as input these vectors are generated from MFCC features computed over the audio using fixed size window length $0.025ms$ in our case these features are fed into seq2seq model (bi-directional RNN) and the final output is validated using CTC loss. Figure adapted from \cite{hannun2014deep} and modified for illustration.}
\label{deepspeech}
\end{figure}
This model was trained on 1.15 hours of original data along with the augmented variations of this data after modifying it to handle attention. The final modified architecture shown in Figure 6. The data was provided by Centre of Speech and Language Technologies institute and is named CSALT which is the only dataset that is phonetically transcribed by expert linguists working in the domain and is publicly available. There are 63 phonemes in Urdu language. \\

\noindent {\fontfamily{qcr}\selectfont  P, P\_H, B, B\_H, M, M\_H, ..., K\_H, ..., L\_H, NG\_H, ..., AE, AEN, A }\\

\noindent In CISAMPA specification these are the representations that are used to represent individual phonemes e.g \#\#D\_D\#\# that gives a heavy 'DH' sound from chest while air is exhaled that is non existent in English language.

\subsubsection{Network Training}
The Connectionist Temporal Classification (CTC) loss is used to train the network where both the input length and output length varies. CTC performs frame to character alignments i.e. a frame's output can span multiple characters which are then merged to compute the final loss, this entails finding the probability of the character $Y$ given the input frame $X$ i.e. $P(Y | X)$. The network was not able to converge using vanilla seq2seq architecture using 1.15 hrs (71.6\% PER). of audio with it's corresponding phonetic representations. The inherent problem of LSTMs in modeling significantly longer sequences also played a role in the poor performance (71.6\% PER) as can be seen from the two empirical examples generated from our experiments in Figure 3  Hyperparameters were SGD optimizer with lr = 0.000001 for all settings.
\begin{figure*}[ht]
\centerline{\includegraphics[width=\textwidth]{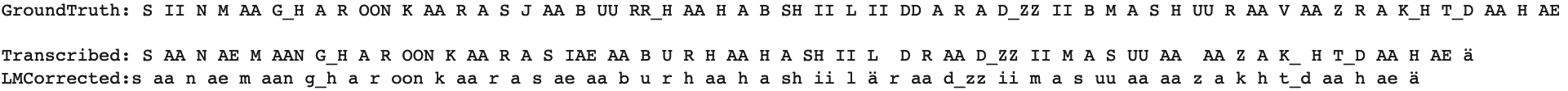}}
\caption{The model struggled to provide good quality phoneme transcriptions over longer sequences even without word boundary identification objective.  }
\label{deepspeech}
\end{figure*}

\subsubsection{Transfer Learning} Due to the problem of non convergence of the network and subpar PER (71.6\%) we decided to apply transfer learning by copying the dense layer weights from earlier layers of a pre-trained English ASR trained on common voice dataset and set the number of output classes from 26 (characters in English language) to 63 (phonemes in Urdu) and reinitialized the last three layers i.e $n, n-1, n-2$ with random weights while keeping the remaining layers weights intact. The layers with transferred weights were frozen for first 10 epochs using $trainable = False$ parameter provided by TensorFlow and were unfrozen starting from 21st epoch. By doing so, our PER improved from $71.6\%$ to $45.2\%$ which provided a significant boost towards convergence. The main intuition behind this approach was that the initial layers of the network are learning the lower level features such as different sounds produced from larynx and vocal tracts i.e. the phonemes, which gets refined as they move forward in the network where towards the end, the word level representations are being formed.

\subsubsection{Attention Mechanism}
Although transfer learning provided significant boost towards convergence, the network was performing poorly in terms of getting the outputs right over longer sequences/audios. The extra objective given to the network was to correctly identify the word boundaries. The network had to pay attention to what it was going to decode next in the local context since a word, its boundaries, and the phonetic sequence of the word lie in the local vicinity. Enter attention mechanism that tries to learn the attention weights such that at a given time step what are the previous and future outputs that need to be considered while producing the phoneme at current time step. The processing over the phonetic sequence can be summarized in equations 1 through 4 as:

\begin{equation}
\vec{h}_t^{(i)}=\vv{LSTM}(f_it), t\in [1,T] 
\end{equation}

\begin{equation}
 \cev{h}_t^{(i)}=\overleftarrow{LSTM}(f_it), t\in [1,T]
 \end{equation}
 
\begin{equation}
 H = [\vec{h}_t;\cev{h}_t] 
 \end{equation}
\begin{equation}
c_t=attention(H[t-6:t+6])
 \end{equation}

\noindent where $f_it$ represents a 25ms frame representing a phoneme in a sequence at time step $t$ and $T$ is the max seq length. After combining the forward and the backward context from BI-LSTM we define the context vector $c_t$ at a certain time step $t$, as $t-6$ to $t+6$ concatenated in order to incorporate the bi-attention width of size $6$ that feeds into the softmax.

We experimented with both self-attention and Bidirectional attention. In order to mimic self attention in bi-directional LSTMs, we concatenated all the context vectors and applied a mask on the future context vectors causing it to essentially act as Masked Attention as proposed in \cite{46201}. For Bidirectional attention no masking was applied to the future phoneme sequences a high level intuition can be seen in the Figure 4.
\begin{figure}[!h]
\centerline{\includegraphics[width=0.5\textwidth]{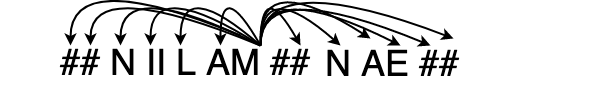}}
\caption{While trying to decode the next phoneme in the sequence, the attention allows the network to pay close attention to the sequence in the local context where the context vector has a width of 6 which enforces locality. In practice the attention is applied to the mel-spectrogram of the 25ms audio frames which corresponds to these sounds. \#\# are the boundaries that the network must learn and are enforced using two pound signs/characters for boundary emphasis.}
\label{attention intuition}
\end{figure}
The modified architecture after adding the attention layer looks like Figure 5 where the embedding vectors in traditional sense are the mel-spectrograms of 25 ms audio. When the network is looking only at the local context, it essentially models that given a sound at a certain time step, what are the other sounds that usually occur around this context. That is, we modeled the context in which certain sounds occur given the current sound and the sounds before and after it. The results for both types of attentions are listed in Table 4.
\begin{figure}
\centerline{\includegraphics[width=0.4\textwidth]{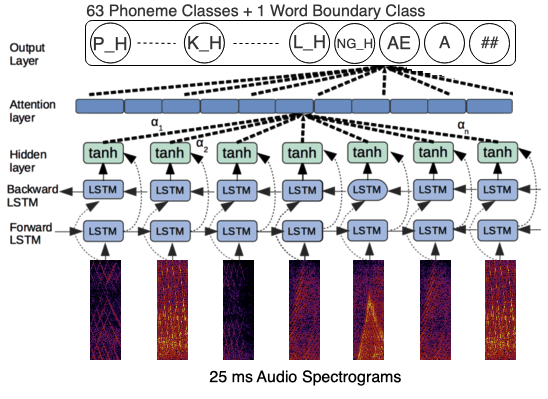}}
\caption{The modified architecture adds an attention layer on top of bidirectional LSTM with a context vector size of 6 as the hyper parameter. selection of context vector in the range 6-15 did not had any significant impact on PER whereas context vector $>$ 15 starts to deteriorate the network performance.}
\label{technical attention}
\end{figure}
\subsubsection{Data Augmentation} In order to overcome data scarcity and better generalization towards different variations of the same phoneme, we generated 10 augmented copies of each audio to get 11.5 hrs. of data in total generated from 1.15 hrs. The augmentations applied to each audio were inspired from SpecAugment \cite{shulby2019robust} and are supported by nlpaug library \cite{ma2019nlpaug}. The used augmentations included: Crop Augment that randomly crops out a section from audio, Loudness Augment that adjusts audio's volume and reduces the volume in certain range at the peak values, Mask Augment that replaces parts of an audio with noise, Noise Augment that injects static noise in the whole audio, Pitch Augment that increases or decreases the pitch, Shift Augment to shift time dimension forward or backward, Speed Augment to increase or decrease the speed, VTLP Augment to change the Vocal Tract length prbutation, Normalize Augment that normalizes the audio. All of the above transforms were applied to the audio with randomization that randomly picks and change random sections of the audio. some examples are shown in Figure 6.
\begin{figure}
\centerline{\includegraphics[width=0.4\textwidth]{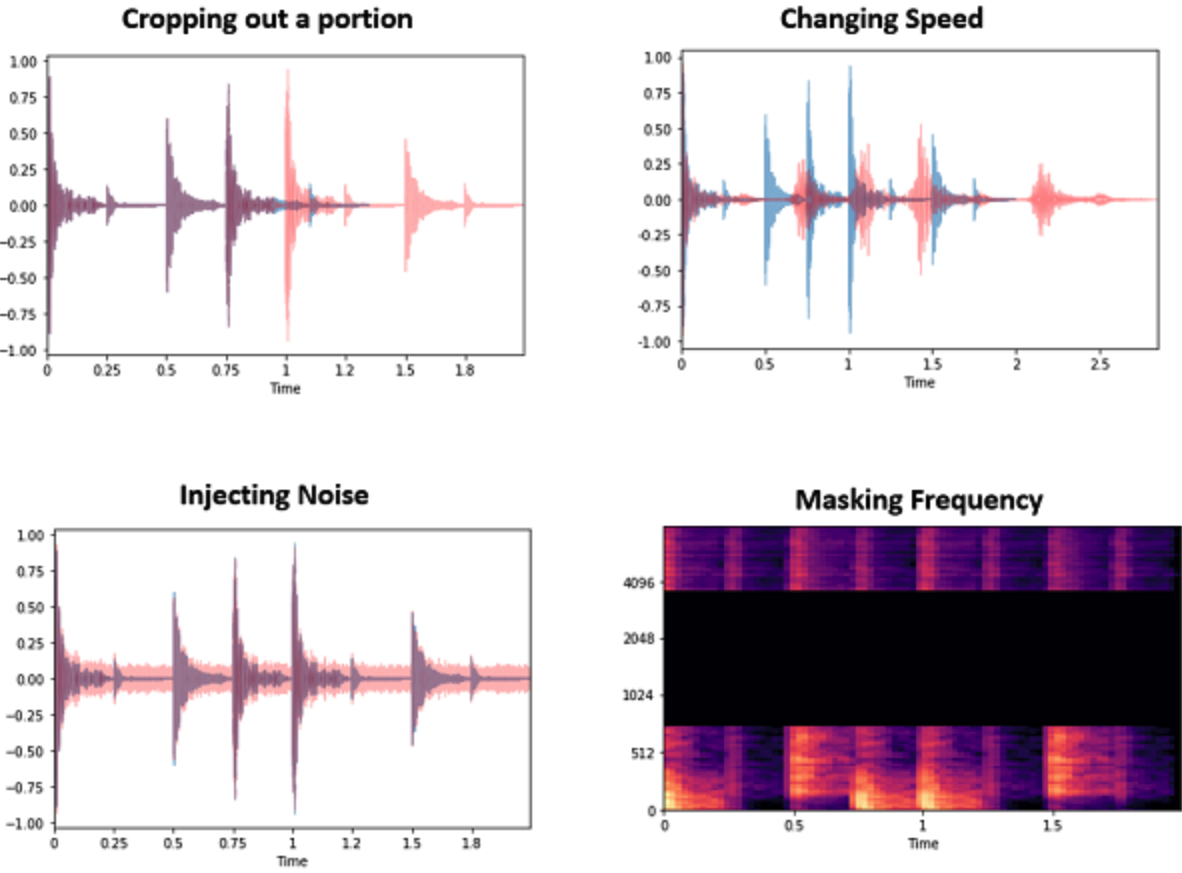}}
\caption{The red sections in the wave form shows the effect of applied transform where as the blue waveform is the original signal. The masking operation is displayed on the spectrogram of the audio signal. A single audio file is taken and Crop Augment, Speed Augment, Noise Augment, and Mask Augment is applied on it and the resulting forms are overlaid on top of the original signal.}
\label{technical attention}
\end{figure}
\section{Evaluation}
There are two parts to our evaluation methodology the first part corresponds to the network convergence using different techniques and the second part corresponds to the semantic evaluation of the network in predicting the word boundaries and phoneme sequences i.e. the Phoneme Error Rate (PER), Word Boundary Accuracy (WBA), and the tri-gram accuracy of the output sequence generated by the network. Before we get into that, we first take a look at the type and amount of data we were working with.
\subsection{Dataset}
The phonetically transcribed data publicly available was 1.15 hrs. which was augmented to 11.5 hrs. containing the same transcriptions as their original counterparts. The dataset had following three parts out of which we only used (1) and (2) for our purpose.
\begin{enumerate}
\item  \textbf{Speech Data} 
\item[]Audio files containing continuous speech with an average of 15 sec per audio and in total was 1.15 hours, made publicly available by Centre of Speech and Language Technologies (CSaLT).

\item \textbf{ Phonetic Transcription:}
The phonetically transcribed data available to us had the word boundaries along with the phonetic representation for each word, there was no temporal information available i.e. the timestamp for each phoneme in the audio. This severely limited us in experimenting with different techniques available out there. The format followed by CSaLT for phoneme transcription is provided below: \\

(\#\#:word bound,$\mathrm{<}$/s$\mathrm{>}$:Sentence Bound)

\item[] {\fontfamily{qcr}\selectfont$\mathrm{<}$s$\mathrm{>}$\#\#N II L A M\#\#N AE\#\#S AA L G I R AA\#\#P A R\#\#H AY DD\#\#S AY S M OO G I R AA F\#\#A S V A D\_D\#\#Q U R AY SH II\#\#K AE\#\#M AA T\_D\_H AE\#\#P A R\#\#AY N TT\_H A N\#\#O R\#\#7 A M\#\#K II\#\#AA T\_D I SH IIN\#\#R O\#\#M E H S UU S\#\#K II\#\#$\mathrm{<}$/s$\mathrm{>}$\ }

\item \textbf{ Text Transcription:}
\item[] \begin{figure}[!h]
\centerline{\includegraphics[width=0.4\textwidth]{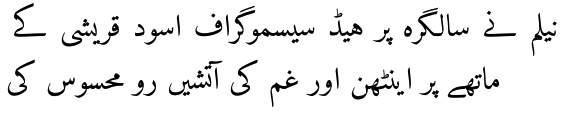}}
\label{kmeans}
\end{figure}
\end{enumerate}
\subsubsection{Network Evaluation}

\noindent The final accuracy for the network was determined using Phoneme Error Rate (PER) where PER is defined as:
\[PER=100\times \frac{I+D+S}{N}\] 
where

\noindent $I$ is the number of phoneme insertions,

\noindent $D$ is the number of phoneme deletions,

\noindent $S$ is the number of phoneme substitutions, and

\noindent $N$ is the total number of phonemes in the target.
\\

\noindent The phoneme error rate before applying transfer learning was $71.6\%$

\noindent For transfer learning we used an English pre-trained model on common voice speech data. We applied transfer learning by changing the network architecture to work for phoneme transcriptions by changing the last layer dimensions and randomly initializing the $n, n-1, n-2$ layers and saw a rapid decline in the network loss which can be seen in Figure 7.

\begin{figure*}[h]
\centerline{\includegraphics[width=0.7\paperwidth]{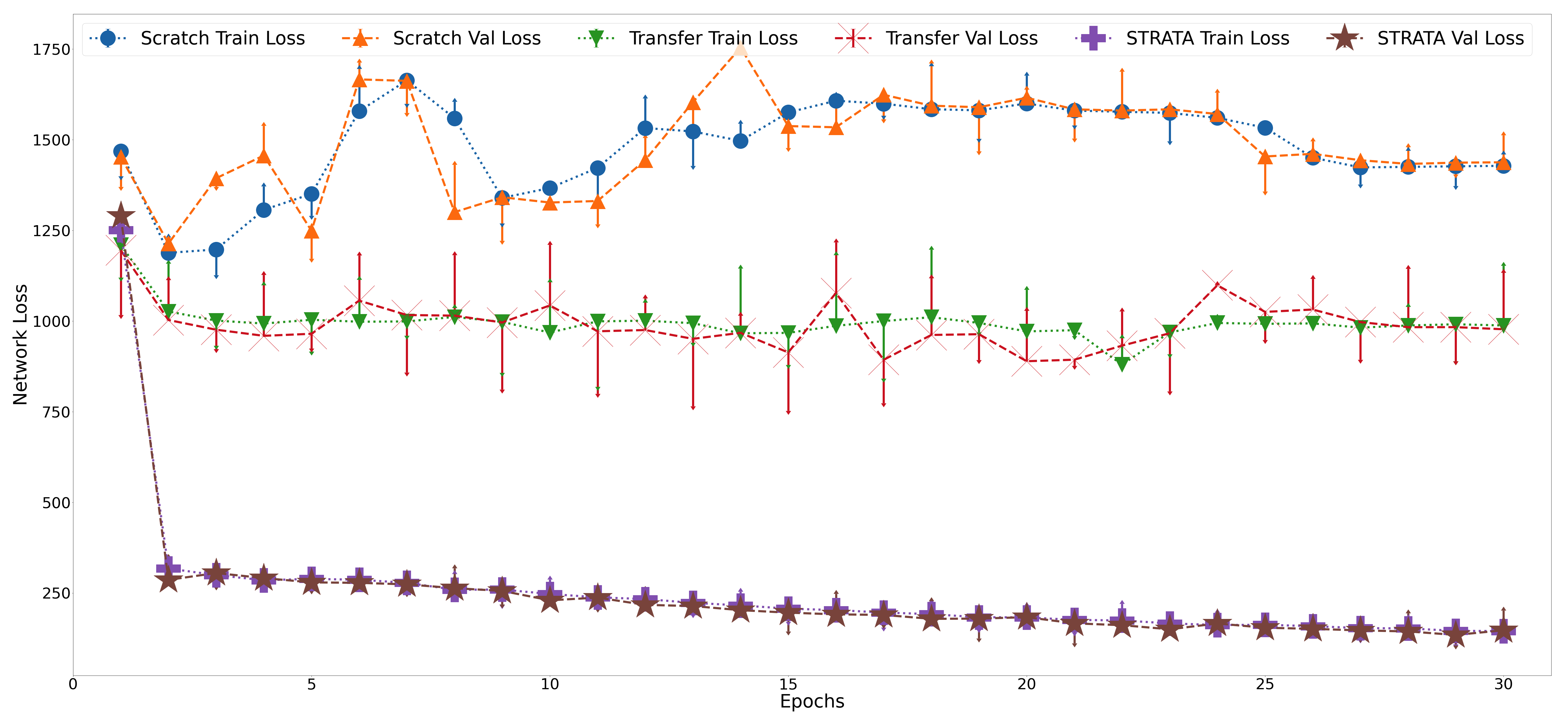}}
\caption{\textbf{Plot with error bounds:} Before transfer learning the loss $(y-\hat{y})$ was above $1600$ and PER was $71.6\%$. The results displayed are averaged over 5 runs of 30 epochs each. The network in the first case struggled to converge and behaved erratically. After transfer learning the loss reduced significantly right from the beginning and the network's Phoneme Error Rate (PER) was $45.2\%$. In this case the network loss was more uniform and did not diverge too much from the mean. Transfer learning along with spec augment and attention mechanism turned out to be the best configuration where the network showed smooth and consistent convergence towards the lowest value of the loss achieved in all our experiments. The significant decline in the network loss starting from first epoch shows that the it was focusing on the phoneme sequence in the local context to get the bigger picture right. Best PER in this case was $24.5\%$ for CSaLT. \\}
\label{main}
\end{figure*}

\begin{figure*}
\centerline{\includegraphics[width=\textwidth]{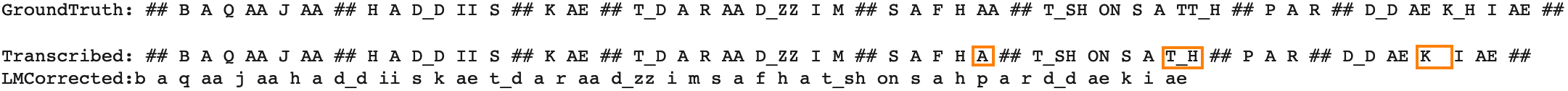}}
\caption{The network trained with \textbf{Attention} provides better outputs for longer sequences even with an added objective of learning to identify word boundaries from continuous speech. The orange boxes represent the only points where the output sequence was different than the ground truth.}
\label{kmeans}
\end{figure*}
\begin{figure}[htb]
\centerline{\includegraphics[width=0.5\textwidth]{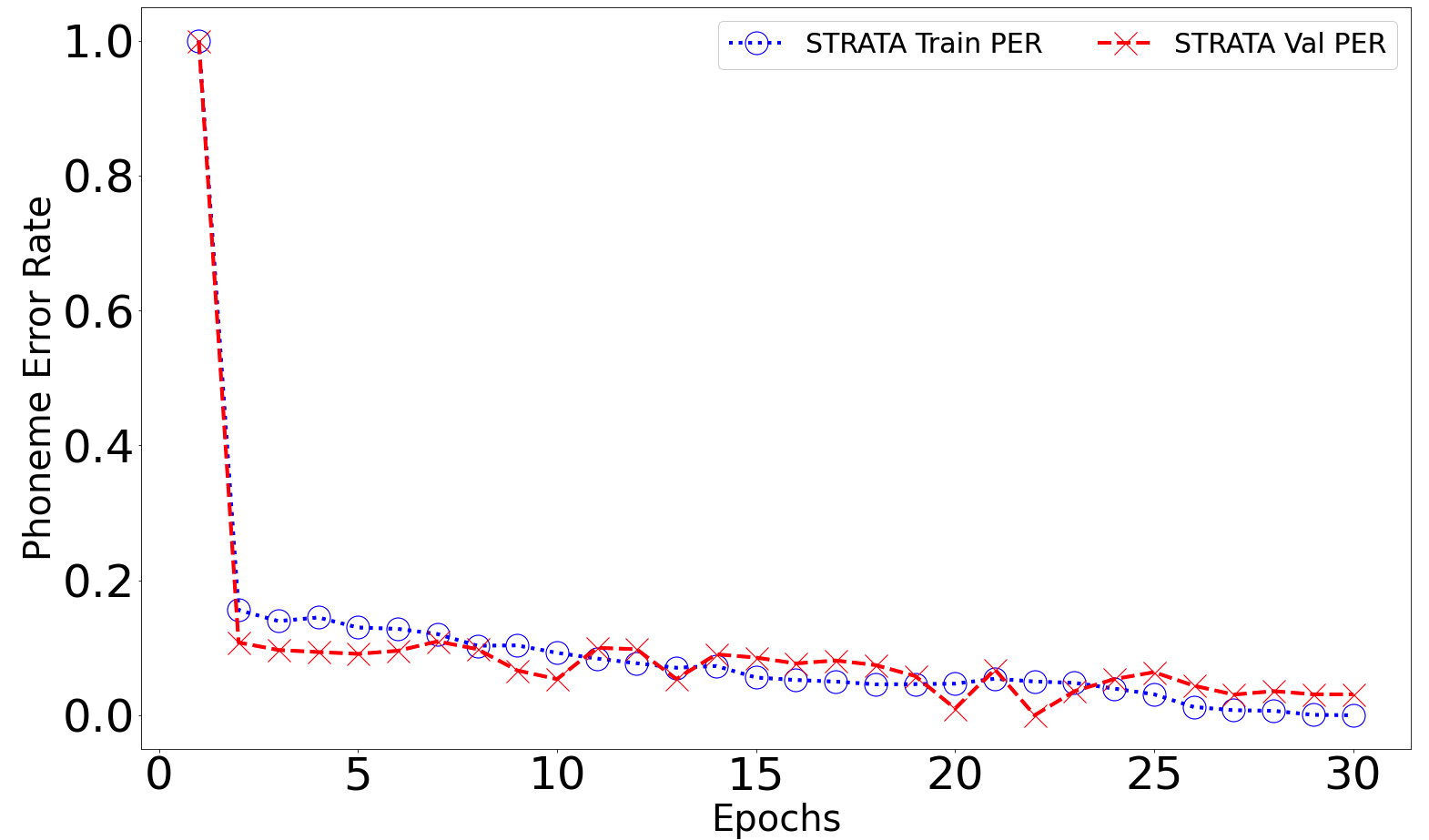}}
\caption{Full configuration: Transfer Learning, Attention, and Data Augmentation our best Phoneme Error Rate (PER) under one of the 5 runs was 16.1\%. But the Average score across 5 runs came out to be 16.5\% for TIMIT dataset. \\}
\label{kmeans}
\end{figure}

\noindent Each training run required approx 4.5 hrs. Using transfer learning saved significant time and resources and the errors were slashed to half in the same amount of time.

\noindent Remember that the pre-trained model had 26 classes corresponding to the characters in English language and was trained on speech to text data. The new model had 64 classes with 63 representing the phonemes and one representing the word boundaries \#\# that was trained for the phonetic representation of the words and their corresponding boundaries (See section 3.1). At each time step the network outputs an index that is mapped to either an individual phoneme in the dictionary e.g \{'1':'P\_H'\} or a word bound at the beginning or ending of word's phonetic sequence e.g \{'64':'\#\#'\}.

\subsubsection{Attention Evaluation}
The results for attention are evaluated for both the Word Boundary Error (WBE) computed in terms of total correct boundaries identified against the ground truth boundaries and the Phoneme Error Rate (PER). The attention results for Bi-Directional attention and self-attention are visualized in Figure 11 \& 12 respectively. \cite{ghaeini2018interpreting}.

\begin{figure*}
\centerline{\includegraphics[width=\textwidth]{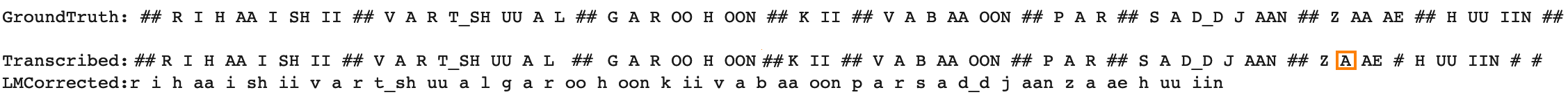}}
\caption{The orange boxes represent the points where the output sequence was different than the ground truth.The effect of using an extra class for identifying word boundaries puts an extra strain on the network to find some pattern that exists between words in a continuous speech. perhaps, an inaudible energy drop that is not perceived by human ears. The net application is that if a boundary separator is emitted by the network we can use it as a signal to append an array of zeros in the original audio to isolate words in continuous speech. \\}
\label{kmeans}
\end{figure*}

\begin{figure}[h]
\centerline{\includegraphics[width=0.5\textwidth]{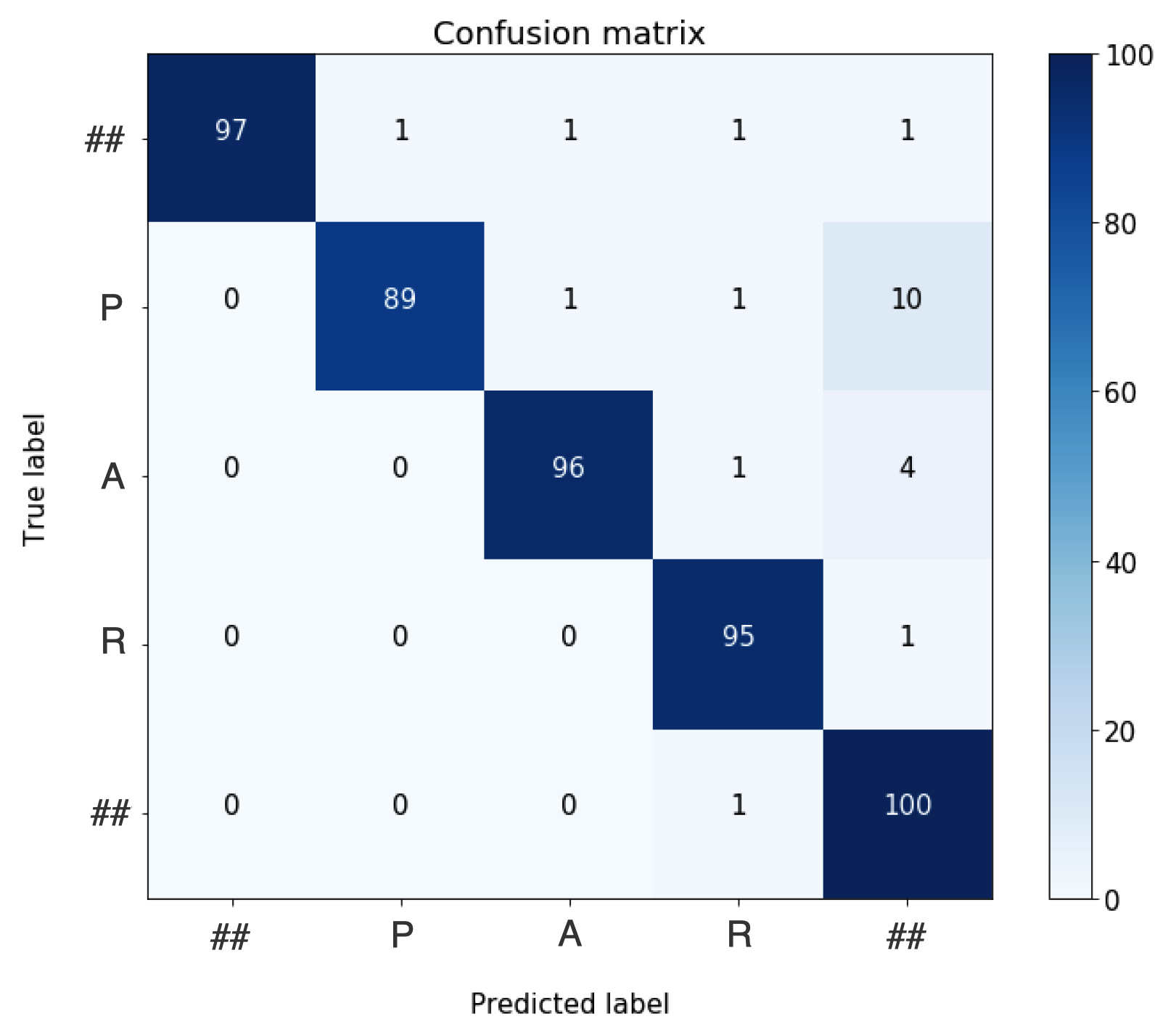}}
\caption{Bi-Directional (Encoder Decoder) Attention weights visualized. Phonetic representation for "Wings" in Urdu. In our experiments the two priors (fwd,bwd) posed by bi-directionality proved to be more effective than self-attention partly in due to low resource setting and having a single prior (bwd) in case of latter. \\ }
\label{kmeans}

\end{figure}

\begin{figure}[!h]
\centerline{\includegraphics[width=0.46\textwidth]{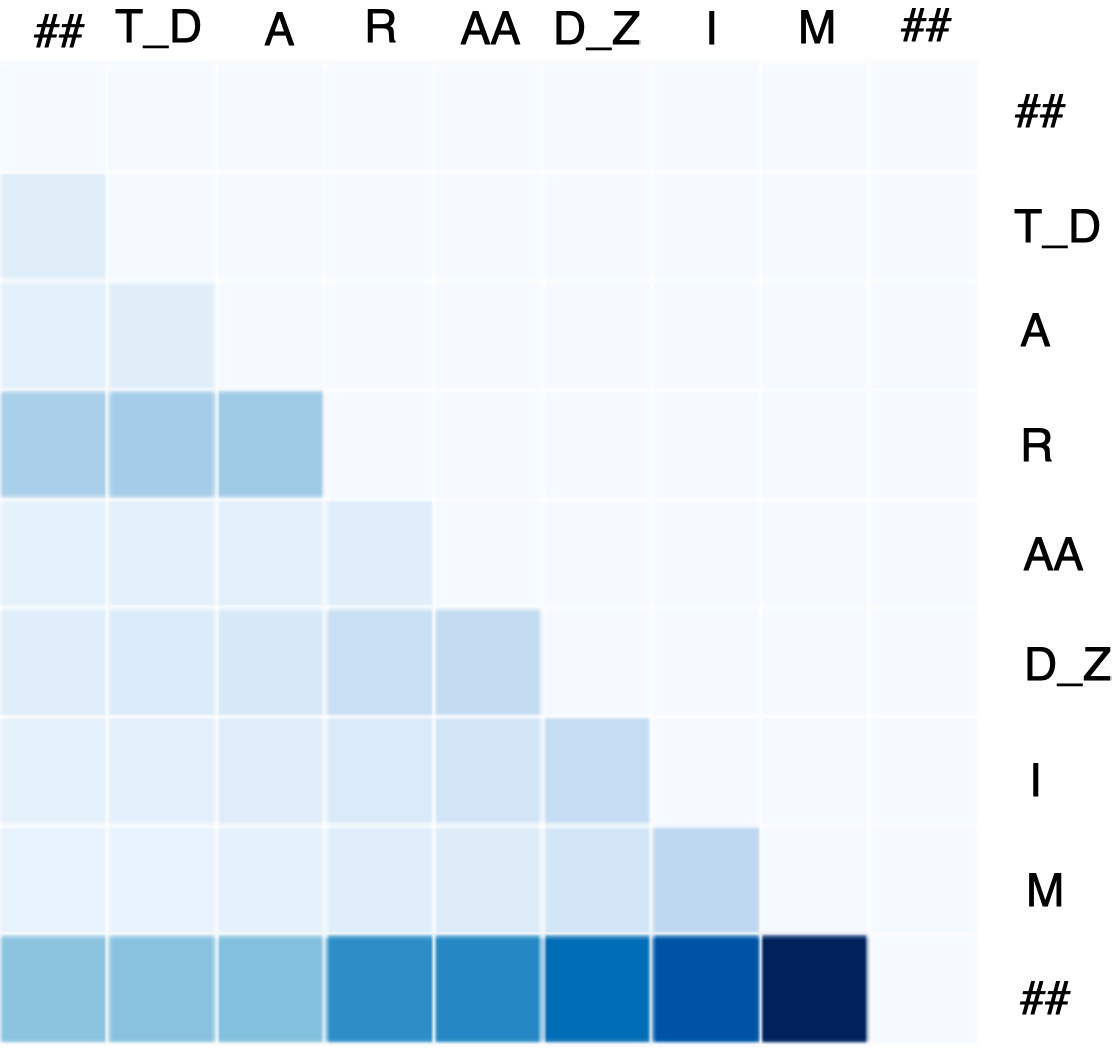}}
\caption{Attention Mechanism was modified for bidirectional LSTMs by adding a mask to the future inputs coming from the backward LSTM. The entries above the diagonal represents the future information and are nullified by adding a matrix containing large negative numbers upper-right that causes softmax to assign zero probability to the future phonemes.}
\label{kmeans}

\end{figure}

Additionally, We computed the tri-gram accuracy for the phoneme sequences generated by the network to evaluate the semantic correctness of the network which tells us how many n-grams in the output transcription appear in the reference ground truth. We evaluated tri-gram accuracy as the count of contiguous sequences of three phonemes output by the network divided by the number of all the tri-grams in the ground truth.

\subsection{Results}
\begin{table}[h]
\centering
\resizebox{0.5\textwidth}{!}{%
\begin{tabular}{ l l l l }
\hline
\textbf{Technique}  & \textbf{PER} & \textbf{Dataset} & \textbf{Year Published} \\ \hline\hline
 GANS (Semi-Supervised) & $43\%$ & TIMIT( 5.4hrs.) & $2019$   \\ \hline
CNN + SVM (Supervised) & $32\%$ & TIMIT( 5.4hrs.) & $2019$    \\ \hline
RNN (Supervised) & $17.7\%$ & TIMIT( 5.4hrs.) & $2013$    \\ \hline
\textbf{
\begin{tabular}[c]{@{}l@{}}STRATA (Supervised) \\ (This Work)\end{tabular}} & \bfseries 16.6\%     & TIMIT( 5.4hrs.)   & $-$                  \\ \hline

\end{tabular}%
}

\caption{\textbf{English Phoneme Recognition Comparison:} STRATA framework when applied to TIMIT dataset was able to improve on the SOTA by 1.1\%. The data augmentation was applied to TIMIT dataset as well which generated an additional 54 hrs. of data. }
\label{tab:my-table}
\end{table}

\begin{table}[h]
\centering
\resizebox{0.5\textwidth}{!}{%
\begin{tabular}{ l l l l }
\hline
\textbf{Technique}  & \textbf{Acc} & \textbf{Dataset} & \textbf{Year Published} \\ \hline\hline
G2P (Supervised) & $64\%$ & CSaLT (1.15hrs.) & $2018$    \\ \hline

\textbf{
\begin{tabular}[c]{@{}l@{}}STRATA (Supervised) \\ (This Work)\end{tabular}} &  75.5\%     & CSaLT (1.15 hrs.)   & $-$                  \\ \hline

\end{tabular}%
}

\caption{\textbf{Urdu Phoneme recognition Comparison:} G2P based model \cite{zia2018pronouncur} gives an accuracy of 64\% using CSaLT text to align words to their phonemes where as our system is able to identify the word boundaries and the phonemes directly from audio speech with an accuracy of 75.5\% for CSaLT dataset (Urdu).}
\label{tab:my-table}
\end{table}

\begin{table}[h]
\centering
\resizebox{0.5\textwidth}{!}{%
\begin{tabular}{ l l l l }
\hline
\textbf{\tiny{Technique}} & \tiny{\textbf{Tri-Gram Acc}}  & \tiny{\textbf{WBA}} \\\hline\hline

\textbf{\begin{tabular}[c]{@{}l@{}}\tiny{Vanilla Seq2Seq} \tiny{(This Work)}\end{tabular}} & \tiny{57.3\%}     & \tiny{67.8\%}      \\ \hline
\textbf{\begin{tabular}[c]{@{}l@{}}\tiny{STRATA - Spec Augment}  \tiny{(This Work)}\end{tabular}} & \tiny{73.2.4\%}     & \tiny{89.7\%}  
 \\ \hline
 \textbf{\begin{tabular}[c]{@{}l@{}}\tiny{STRATA} \tiny{(This Work)}\end{tabular}} & \tiny{\bfseries 87.3\%}     & \tiny{\bfseries 92.2\%} 
  \\ \hline
\end{tabular}%

}
\caption{\textbf{Seq2Seq:} Tri-Gram accuracy  taken against Ground Truth, and Word Boundary Accuracy (WBA) defined in terms of the correctly identified boundaries in the output that were also there in the ground truth.}

\label{tab:my-table}
\end{table}
\begin{table}[h]
\centering
\resizebox{0.5\textwidth}{!}{%
\begin{tabular}{ P{3cm} P{2cm} }
\hline
\textbf{\tiny{Attention}} & \textbf{\tiny{PER}}  \\ 
\hline\hline
\tiny{\textbf{STRATA (Self-Attention)}} & \tiny{ 29.8\%} \\ 
\hline
\tiny{\textbf{STRATA (Bi-Attention)}} & \tiny{\bfseries 24.5\%}  \\ 
\hline
\end{tabular}%
}

\caption{\textbf{Attention Comparison:} Bidirectional-Attention which is analogous to Encoder-Decoder attention in transformers performed much better for low resource setting (CSaLT)  where both the forward context and the backward context is taken into consideration while producing the next output when compared to Self-Attention.}
\label{tab:my-table}
\end{table}
\noindent The Tri-Gram accuracy in phoneme recognition is calculated by using the tri-grams in the original phonetic transcription as the reference text where as the tri-grams emitted by the trained model are treated as the predicted sequence. The WBA is computed as the words whose boundaries were correctly identified by the network in the phonetic transcripts over the total ground truth word boundaries. WBA is a dumb metric in that it does not factor in the semantics of the transcriptions due to its indifference towards the context of the phonemes or what is happening between the bounds i.e it does not care about whether "P\_H" came before "AE" or after it. The N-Gram based accuracy score is a bit more intelligent in that sense as it was computed for the tri-grams i.e. a lower score is assigned when the sequence of occurrence in the correct order is not observed including the word bounds. We further hypothesize that the success of data augmentation at phonetic level was largely in part due to the fact that the different variations of each  basic sound unit were captured from the limited dataset. 
\section{Related Work}
\noindent Research in Phoneme recognition is still an active area of research and is considered as not a fully solved problem in computational linguistics especially in languages other than English where no standardized dataset such as TIMIT exists. Phoneme recognition from audio can be categorized into three broad categories supervised, semi-supervised, and hybrid. 

\noindent \textbf{Supervised} techniques includes HMM (Hidden Markov Models) which is a supervised phoneme modeling technique containing individual states (Finite Automata) representing phonemes and the edges between these states represent the transition probabilities from one state to another \citep{mohri2008speech} where the transitions from one state to another are strictly defined by their previous states. Another technique that was used to recognize phonemes employed SVM (Support vector Machines) in order to classify the audio signal into one of the 40 phonemes based on multi class setting as defined by \citep{crammer2001algorithmic} and reported a 32\% PER. Convolutional Neural Networks using spectral features of the frame were also used in some cases for phoneme recognition task \citep{glackin2018convolutional}. Recurrent Neural networks and Long-Term Short memory networks \citep{graves2005bidirectional} are useful in modeling the time element of the phoneme sequence while implicitly learning the phoneme transition probabilities. The RNN based approach used by \cite{graves2013speech} treated phoneme recognition as a sequence labeling problem and were able to achieve 17.7\% Phoneme Error Rate.

\noindent \textbf{Semi-Supervised} approaches include training a Recurrent Neural Network on labeled subset of the data which is then used to label the rest of the data \citep{tietz2017semi} and finally the model is trained on complete data. This approach suffers from the quality of labels that are assigned to the unlabeled data. GANs have been used with generator discriminator concept \citep{chen2019completely} where generator network generates phoneme labels corresponding to their acoustic features and the discriminator module having the correct label knowledge is used to calculate the loss. Although the authors claimed it to be unsupervised but the fact that there is a discriminator that has the correct knowledge of the actual transcripts and is used to determine the goodness of the labels is still considered to be semi-supervised.

\noindent \textbf{Hybrid} approaches have become popular in recent years where the common theme is to apply deep neural networks along with non neural network based classifiers such as SVM, HMM e.t.c \citep{graves2005bidirectional} performed frame-wise phoneme classification  using BLSTM-HMM based model where the Bidirectional LSTMs were used as estimator of acoustic probabilities and the parameters of HMM were estimated via Viterbi Training. They  carried out the experiments on TIMIT dataset containing a lexicon of 61 phonemes and reported the classification score of 86.4\%.
Another approach followed by \citep{shulby2019robust} consisted of segmenting the audio spectrograms according to the phoneme transcriptions that consisted the information about each frame and its corresponding duration. They extracted the features from these spectrograms using convolutional neural networks followed by classification via Support Vector Machines and reported 32\% PER for their best setting. \\


\section{Conclusion}
In this work we demonstrated an approach for tackling the data scarcity problem using Transfer Learning, Attention Mechanism, and Data Augmentation for extracting Word Boundaries and the phonetic representations of those words from continuous audio speech for an under resourced language like Urdu. The CSaLT dataset used for Urdu is also relatively easier to construct than TIMIT dataset and can be replicated easily for other low resource languages.
Our approach produced better results as compared to other supervised techniques used for phoneme recognition. To the best of our knowledge, this approach haven't been used to accomplish this task before for under resourced languages. Transfer learning combined with attention mechanism proved to be effective in significantly reducing the training time and error rates and provided  significant gains with accuracy. Finally, the data augmentation provided us with much needed variation in the sounds that made it possible for the network to generalize better. The word boundary objective used in our approach can be used to isolate the words in continuous speech that can be useful in keyword detection and trigger word detection applications. We also plan to train an End to End ASR using STRATA  for low resource languages like Urdu in our future work.



\bibliography{anthology}
\bibliographystyle{acl_natbib}

\end{document}